\def\year{2022}\relax
\documentclass[letterpaper]{article} 
\usepackage{aaai22} 
\nocopyright
\usepackage{times} 
\usepackage{helvet} 
\usepackage{courier} 
\usepackage[hyphens]{url} 
\usepackage{graphicx} 
\usepackage{natbib} 
\usepackage{amsmath}
\usepackage{caption} 
\DeclareCaptionStyle{ruled}{labelfont=normalfont,labelsep=colon,strut=off} 
\usepackage{algorithm}
\usepackage{algorithmic}
\usepackage{amssymb}
\usepackage{bm}
\usepackage{siunitx}

\usepackage{todonotes}

\usepackage{newfloat}
\usepackage{listings}
\floatstyle{ruled}
\newfloat{listing}{tb}{lst}{}
\floatname{listing}{Listing}
\usepackage{booktabs}
\usepackage{adjustbox}

\usepackage{caption}
\usepackage{subcaption}

\newcommand{\mname}{\texttt{DRG-LLaMA} }

\title{\mname: Tuning LLaMA Model to Predict Diagnosis-related Group for Hospitalized Patients}
\author {
  Hanyin Wang,\textsuperscript{\rm 1}
  Chufan Gao,\textsuperscript{\rm 2} 
  Christopher Dantona,\textsuperscript{\rm 3} 
  Bryan Hull,\textsuperscript{\rm 4}
  Jimeng Sun\textsuperscript{\rm 2,5} 
}
\affiliations {
  \textsuperscript{\rm 1} Division of Hospital Internal Medicine, Mayo Clinic Health System, Mankato, Minnesota\\
  \textsuperscript{\rm 2} Department of Computer Science, University of Illinois Urbana-Champaign, Champaign, Illinois\\
  \textsuperscript{\rm 3} Enterprise Inpatient Clinical Documentation Integrity, Mayo Clinic, Rochester, Minnesota\\
    \textsuperscript{\rm 4} Division of Hospital Internal Medicine, Mayo Clinic, Phoenix, Arizona\\
  \textsuperscript{\rm 5} Carle Illinois College of Medicine, University of Illinois Urbana-Champaign\\
  wang.hanyin@mayo.edu
}

\usepackage{bibentry}

\begin{document}

\maketitle

\section{Abstract}
In the U.S. inpatient payment system, the Diagnosis-Related Group (DRG) is pivotal, but its assignment process is inefficient. The study introduces \mname, an advanced large language model (LLM) fine-tuned on clinical notes to enhance DRGs assignment. Utilizing LLaMA as the foundational model and optimizing it through Low-Rank Adaptation (LoRA) on 236,192 MIMIC-IV discharge summaries, our \mname-7B model exhibited a noteworthy macro-averaged F1 score of 0.327, a top-1 prediction accuracy of 52.0\%, and a macro-averaged Area Under the Curve (AUC) of 0.986, with a maximum input token length of 512. This model surpassed the performance of prior leading models in DRG prediction, showing a relative improvement of 40.3\% and 35.7\% in macro-averaged F1 score compared to ClinicalBERT and CAML, respectively. Applied to base DRG and complication or comorbidity (CC)/major complication or comorbidity (MCC) prediction, \mname achieved a top-1 prediction accuracy of 67.8\% and 67.5\%, respectively. Additionally, our findings indicate that \mname’s performance correlates with increased model parameters and input context lengths.

\section{Introduction}
The emergence of LLMs, such as GPT-3 \cite{brown2020language} and InstructGPT \cite{ouyang2022training}, has brought about a transformative shift in the landscape of Natural Language Processing (NLP). These LLMs have demonstrated exceptional capabilities across many NLP tasks in the general domain. However, the integration of LLMs into the medical field remains at a nascent stage within the academic community. Recent instances of progress highlight their significant potential, including OpenAI's GPT-4 \cite{nori2023capabilities}, Google's Med-PaLM2 \cite{singhal2023expertlevel}, and Google Deepmind's Med-PaLM M \cite{tu2023towards}. GPT-4 and Med-PaLM 2 have achieved impressive performance on the United States Medical Licensing Examination (USMLE), and Med-PaLM M can even classify radiology images. 
Nonetheless, the medical domain introduces elevated concerns regarding safety and privacy, necessitating detailed analysis regarding the performance and limitations of LLMs to address the inherent risks such as hallucination, bias, and reasoning deficiencies \cite{au2023ai}. 

Since its inception by Medicare in 1983, DRG has served as the foundation for the inpatient prospective payment system within the United States \cite{quinn2014after}. 
Each distinct DRG code is delineated by a particular set of patient attributes, including principal diagnosis, specific secondary diagnoses, procedures, sex and discharge status \cite{centers2016icd}. 
Traditionally, the assignment of DRGs constitutes a labor-intensive manual endeavor undertaken by coding specialists, typically subsequent to a patient's discharge. 
Given the pivotal role of DRGs and their bundled metrics (e.g., case-mix index, geometric length of stay) in the operational and financial performance of hospitals, a pressing interest exists in the accurate early prediction of DRGs during a patient's hospitalization. This prediction is vital for efficacious resource planning and allocation.
The task of DRG prediction presents distinct challenges compared to automated International Classification of Diseases (ICD) coding. This distinction stems from differences in the nature of the task: DRGs involve multi-class classification, where one DRG code is assigned to each visit, in contrast to the multi-label classification of ICDs, where multiple codes may apply to a single visit \cite{kaur2022ai}. Additionally, the hierarchical structure of the codes, such as the presence of a principal diagnosis in DRGs, and the context of utilization in hospital operations further differentiate the two tasks \cite{centers2016icd}.
Previous studies have showcased advancements in DRGs classification accuracy through various machine-learning algorithms \cite{gartner2015machine} and deep neural networks \cite{islam2021deepdrg}. 
More recently, a deep learning-based NLP model leveraging adjusted Convolutional Attention for Multi-Label Classification (CAML) has been applied to predict DRGs based on clinical notes and yielded promising outcomes \cite{mullenbach2018explainable, liu2021early}.

With LLM's remarkable natural language synthesis and generating capabilities, we hypothesize LLM could be applied to effectively predict DRGs directly from clinical notes. 
In this work, we present \mname, a fine-tuned LLM derived from LLaMA \cite{touvron2023llama}.
\mname is trained on discharge summaries from the MIMIC-IV dataset for the task of DRG prediction. 
In our investigation, we approached DRG prediction from two perspectives: 1) as a single-label classification task, where the model makes an end-to-end prediction of the DRG label, and 2) as a two-label classification task, where the model predicts base DRG and CC/MCC status as two separate labels, followed by the inference of the final DRG label from these two components (i.e., base DRG and CC/MCC status). Our work revealed superior performance of \mname in DRG prediction compared to the previously reported leading models of CAML \cite{liu2021early} and ClinicalBERT \cite{alsentzer-etal-2019-publicly}.

\section{Results} 

\subsection{Study cohort}
\begin{figure}[t]
\includegraphics[width=0.48\textwidth]{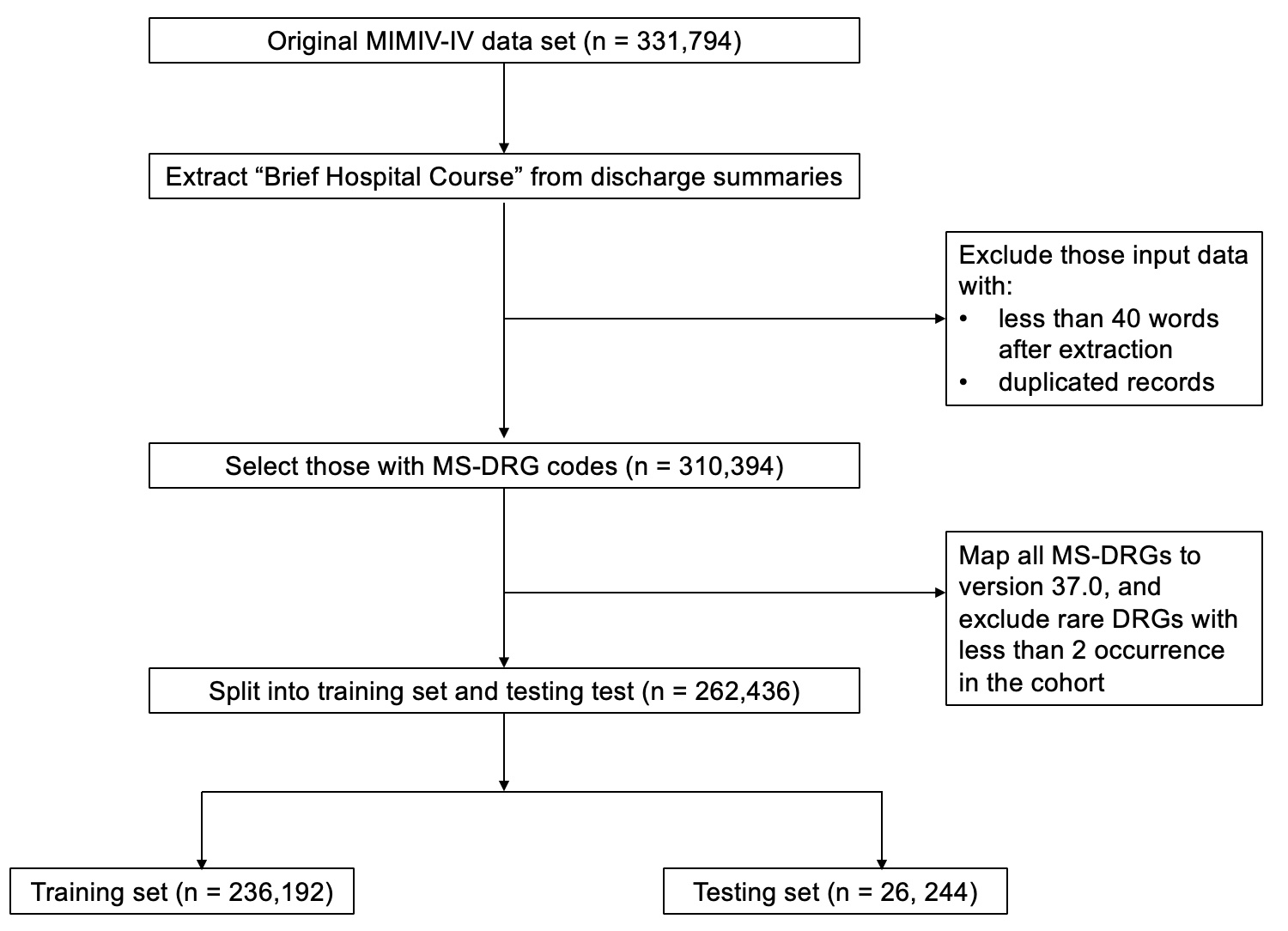}
\captionsetup{font={bf}}
\caption{Flow diagram of the cohort processing steps.}
\centering
\label{Figure:Flow diagram}
\end{figure}

A summary of the study cohort and data preprocessing steps was shown in Figure \ref{Figure:Flow diagram}. 
We focused on hospital stays with Medicare severity-DRGs (MS-DRGs) within the MIMIC-IV dataset. 
The ``brief hospital course'' section from discharge summary was extracted to serve as input text. 
We also filtered out low-quality discharge summaries and rare DRGs with less than 2 occurences in the cohort. 
90\% of the data was allocated as training set while the rest 10\% as testing set, and this partitioning was stratified on DRGs. 
The training and testing set contains 738 and 723 unique DRG labels, respectively.
There is no significant difference in the average word counts in the training vs. testing set (398 vs. 399; p = 0.51 from two-sided t-test). The distribution of cases per DRG is imbalanced, with a median number of 124.5 in the training set (Supplementary Figure 1).

\subsection{DRG prediction as a single-label classification task}
We presented the results with a maximum input token size of 512 in Table \ref{Table:single label}. 
\mname consistently outperformed ClinicalBERT and CAML across all evaluation metrics, with the most notable contrast seen in macro-F1 score (showing a relative improvement of 40.3\% and 35.7\% compared to ClinicalBERT and CAML, respectively).  
The accuracy of top-1 and top-5 predictions achieved by our fine-tuned \mname-7B model was 52.0\% and 84.8\%, respectively. When only considering the most frequent 300 DRGs, the top-1 accuracy improved to 55.7\%, and this further increased to 69.4\% in the most frequent 30 DRGs. As expected, \mname's performance declined in less frequent DRGs (Figure \ref{fig:2a}).
When compared to CAML, ClinicalBERT achieved higher AUC and top-1 prediction accuracy but lower macro-averaged F1 score. High AUC scores were obtained for all models due to the many infrequent DRG classes, resulting in high true negative predictions for all negative class predictions. \cite{liu2021early}. 

We investigated \mname's performance across varying model sizes and input context lengths (Table \ref{Table:different model sizes}), observing a consistent improvement in all evaluation metrics with larger models and longer input contexts, measured in maximum token numbers. The optimal configuration, utilizing a 13B LLaMA model and a maximum input token size of 1024, achieved a top-1 prediction accuracy of 54.6\%, a top-5 prediction accuracy of 86.5\%, and a macro-F1 score of 0.361.

\begin{table*}[!ht]
\small
\captionsetup{font={bf}}
\caption{Main Results on DRG prediction with a max input token size of 512}
\label{Table:single label}
 \begin{tabular}{l l p{1cm} p{1.5cm}p{1.5cm}p{1.5cm} p{1.5cm} p{1.5cm} p{1.5cm}} 
 \toprule
Model & DRG set  & MACRO-F1 & ACC@1 & ACC@5 & ACC@10 & MACRO-AUC & MICRO-AUC & Number (\%) of cases \\ 
\midrule
   \mname-7B & All DRGs & \textbf{0.327 (0.004) } &\textbf{0.520 (0.003)} &\textbf{0.848 (0.002)} &\textbf{0.912 (0.002)} & \textbf{0.986 (0.001)} & \textbf{0.994 (0.000) } & 26,244 (100.0)\\ 
    & Top 300 DRGs & 0.497 (0.005) & 0.557 (0.004) & 0.876 (0.002) & 0.932 (0.001) & 0.988 (0.000) & 0.995 (0.000) & 22,940 (87.4)\\ 
    & Top 50 DRGs  & 0.700 (0.004) & 0.666 (0.004) & 0.931 (0.002) & 0.965 (0.001) & 0.989 (0.000) & 0.998 (0.000) & 10,270 (39.1) \\
    & Top 30 DRGs  & 0.737 (0.005) & 0.694 (0.005) & 0.941 (0.003) & 0.971 (0.002) & 0.988 (0.001) & 0.998 (0.000) & 7,666 (29.2)\\\\
   ClinicalBERT & All DRGs  & 0.233 (0.003) & 0.502 (0.003) & 0.815 (0.002) & 0.881 (0.002) & 0.979 (0.001) & 0.991 (0.000) & 26,244 (100.0)\\\\ 
   CAML & All DRGs  & 0.241 (0.003) & 0.447 (0.002) & 0.785 (0.002) & 0.865 (0.002) & 0.976 (0.001) & 0.991 (0.000) & 26,244 (100.0) \\
 \bottomrule
\end{tabular}
\\
F1 and AUC scores were calculated using macro-averaged or micro-averaged method as shown in the header. Notably, in a multi-class classification problem, micro-averaged F1 score is equal to top-1 prediction accuracy when labels of all classes are considered \cite{grandini2020metrics}.  Accuracy @1, @5 and @10 measure whether the top-1, top-5 and top-10 predictions by the model contain correct DRG code, respectively. Standard deviations are shown in parentheses and calculated using a bootstrapping procedure. Top DRGs are selected based on the number of cases per DRG in the dataset. Number (\%) of cases represents hospital stays covered by the given DRG group in the testing set. \textbf{Bolded scores} denote the best performance with respect to the task. \mname outperformed ClinicalBERT and CAML across all evaluation metrics,  with better performance in more frequent DRGs. \textit{DRG} denotes diagnostis-related group, \textit{AUC} denotes area under the receiver operating characteristic curve, and \textit{ACC} denotes accuracy. 
\end{table*}

\begin{table*}[!ht]
\small\centering
\captionsetup{font={bf}}
\caption{\mname performance on different model and max input token sizes}
\label{Table:different model sizes}
 \begin{tabular}{l p{2.2cm} p{1.5cm} p{1.5cm}p{1.5cm}p{1.5cm} p{1.5cm} p{1.5cm} } 
 \toprule
Model size & Max input token size & MACRO-F1 & ACC@1 & ACC@5 & ACC@10 & MACRO-AUC & MICRO-AUC \\ 
\midrule
   13B & 1024 & \textbf{0.361 (0.004)} & \textbf{0.546 (0.003)} &  \textbf{0.865 (0.002)} & \textbf{0.925 (0.001)} & \textbf{0.986 (0.001)} & \textbf{0.994 (0.000)}\\ 
    & 512 & 0.334 (0.005) & 0.524 (0.002) & 0.853 (0.002) & 0.914 (0.002) & 0.984 (0.001) & 0.993 (0.000) \\
    & 340 & 0.312 (0.006) & 0.499 (0.003) & 0.834 (0.002) & 0.902 (0.002) & 0.983 (0.001) & 0.992 (0.000)\\
   
   7B  & 1024  & 0.346 (0.004)  & 0.539 (0.003) & 0.861 (0.002) & 0.923 (0.001) & 0.986 (0.001) & 0.994 (0.000) \\
    & 512 & 0.327 (0.004) & 0.520 (0.003) & 0.848 (0.002) & 0.912 (0.002) & 0.986 (0.001) & 0.994 (0.000)  \\ 
    & 340  & 0.303 (0.005) & 0.493 (0.003)  & 0.828 (0.002) & 0.896 (0.002) & 0.981 (0.001) & 0.992 (0.001) \\

 \bottomrule
 \end{tabular}
 
 Experiments were performed on LLaMA with a size of 7 billion and 13 billion parameters. \textbf{Bolded scores} denote the best performance. We observed that \mname’s performance consistently improved with larger models and longer input contexts.  
\end{table*}

\subsection{DRG prediction as a two-label classification task}
\label{sec:two_label}
In the two-label approach, we first dissect each DRG into two distinct components: a base DRG label and a CC/MCC label (denoting complication or comorbidity / major complication or comorbidity). This dissection process was based on the composition delineated within the MS-DRG v34.0 definitions manual \cite{centers2016icd}. 
The five distinct labels attributed to CC/MCC are as follows: ``without CC/MCC'', ``with CC'', ``with MCC'', ``without MCC'', and ``not applicable''. 
As an example, in DRG code 53 of "spinal disorders and injuries without CC/MCC," "spinal disorders and injuries" represents the base DRG label, while "without CC/MCC" serves as the CC/MCC label. Following this mapping process, the 738 DRG codes were converted into a combination of 340 base DRG labels each paired with one of the five CC/MCC labels.
Results of 
two-label approach using \mname-7B with a maximum input token size of 512 was shown in Table \ref{Table:two label}. 
The top-1 prediction accuracy for base DRG and CC/MCC reached 67.8\% and 67.5\% respectively. 
This result suggests that predicting the principal diagnosis or procedure without considering CC/MCC is a significantly easier task on its own.

Upon integrating a mapping rule designed to infer DRGs through the combination of base DRG and CC/MCC labels, the accuracy reached 51.5\% across all DRGs. 
Notably, this performance was comparable with the accuracy attained in the single-label approach of 52.0\% using the same base model, showing that the LLM was able to achieve state-of-the-art performance via either classification setting. 

\begin{table*}[!ht]
\small\centering
\captionsetup{font={bf}}
\caption{Main Results on DRG prediction as a two-label task with a max input token size of 512}
\label{Table:two label}
 \begin{tabular}{l p{1.5cm} p{1.5cm}p{1.5cm}p{1.5cm} p{1.5cm} p{1.5cm}  p{1.5cm} } 
 \toprule
Component & MACRO-F1 & ACC@1 & ACC@5 & ACC@10 & MACRO-AUC & MICRO-AUC & Number of labels\\ 
\midrule
   Base DRG  & 0.520 (0.005) & 0.678 (0.002) & 0.912 (0.001) & 0.953 (0.001) & 0.990 (0.001)& 0.995 (0.000) & 340\\
   CC/MCC  & 0.680 (0.003) & 0.675 (0.003) & - & - & 0.909 (0.001)& 0.918 (0.001) & 5\\
   DRG & - &  0.515 (0.003) & - & - & - & - & 738\\
 \bottomrule
 \end{tabular}

Experiments were performed with \mname-7B and a maximum input token size of 512. The top-1 prediction accuracy for base DRG and CC/MCC reached 67.8\% and 67.5\% respectively. A top-1 prediction accuracy of 51.5\% was achieved by employing the mapping rule on base DRG and CC/MCC labels, as elaborated in the methodology section. \hspace{9cm}
\end{table*}

\subsection{Error analysis}
\begin{figure}[p]
     \centering
     \captionsetup{labelfont=bf={bf}}
     \begin{subfigure}[b]{0.5\textwidth}
         \centering
         \includegraphics[width=\textwidth]{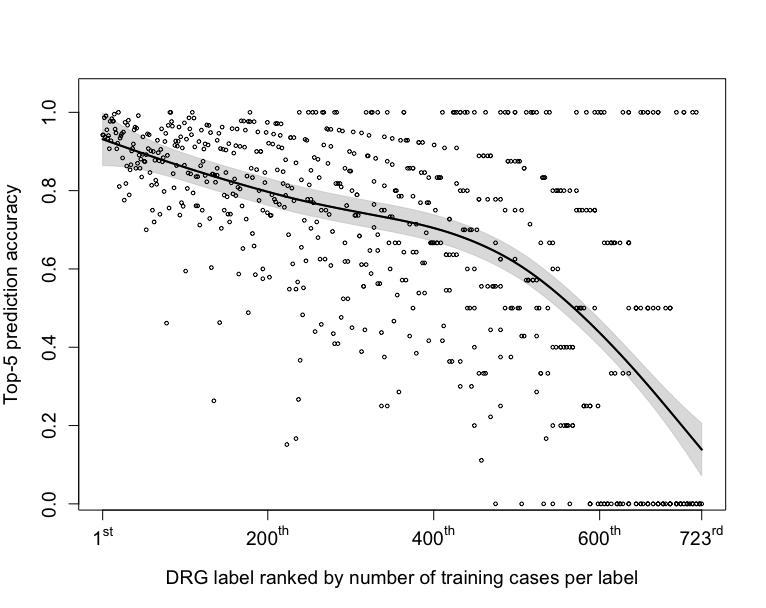}
         \caption{}
         \label{fig:2a}
     \end{subfigure}
     \begin{subfigure}[b]{0.5\textwidth}
         \centering
         \includegraphics[width=\textwidth]{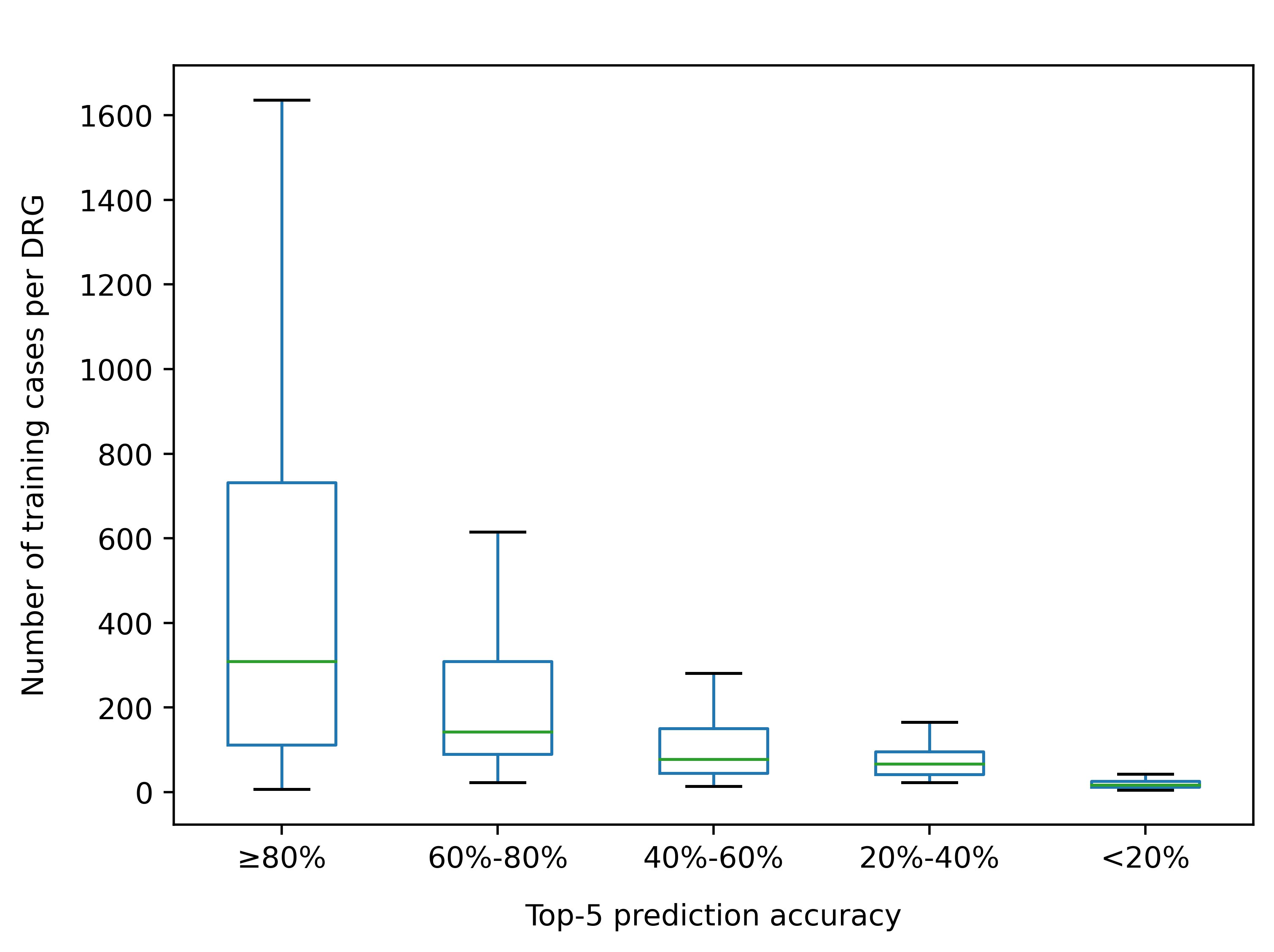}
         \caption{}
         \label{fig:2b}
     \end{subfigure}
        \caption{\textbf{Relationship between training cases per DRG and prediction accuracy by \mname.} Results from \mname-7B with a maximum input token size of 512. \textbf{(a)} Scatter plot of top-5 prediction accuracy versus DRG ranks by number of training cases. Y-axis is top-5 prediction accuracy of each DRG label. X-axis is the rank of the 723 DRGs by their number of training cases, where DRG ranked 1\textsuperscript{st} has the most training cases, and DRG ranked 723\textsuperscript{rd} has the least training cases. Black dots indicate individual DRGs. The solid line represents smoothing spline estimated relationship (equivalent degrees of freedom: 6.35; R\textsuperscript{2}: 0.434). The gray shaded area denotes a 95\% Bayesian confidence interval for the smoothing spline estimated function. As expected, \mname's performance declined in less frequent DRGs. \textbf{(b)} Boxplot of training cases per DRG with groups of different prediction accuracy. DRGs are grouped by range of top-5 prediction accuracy as shown in X-axis. Y-axis is the number of training cases per DRG. The green line represents the median value; the box limits show the interquartile range (IQR) from the first (Q1) to third (Q3) quartiles; the whiskers extend to the furthest data point within Q1–1.5*IQR (bottom) and Q3+1.5*IQR (top). DRG groups with better prediction performance generally have a greater number of training cases, although there is a large variance in the number of training cases within the best performing group.}
\end{figure}

As noted above, a correlation exists between the number of training cases and prediction performance. The accuracy of DRG prediction depends on various factors. DRGs with a top-5 prediction accuracy exceeding 80\% are typically associated with a median of 309 training cases per label. In contrast, those DRGs with a top-5 accuracy below 20\% are associated with only a median of 17 training cases per label (as shown in Figure \ref{fig:2b}). However, other factors, such as the type of DRG, also affect prediction performance. For instance, out of the DRGs with a top-1 prediction accuracy of 100\%, 8 out of 9 are surgical DRGs, which have distinct hospital courses that make them easier for the model to comprehend (as listed in Supplementary Table 2). We randomly selected 10 samples from the subset where the model presented erroneous predictions within its top ten outcomes for manual error analysis (as listed in Table \ref{Table:error analysis}).
Broadly, the identified errors were categorized as follows: erroneous CC/MCC (1/10), correct information needed for DRG prediction unavailable (1/10), difficulty in selecting correct base DRG (3/10), inadequate clinical concept extraction (4/10) and an isolated case of a plausible incorrect DRG label (1/10). Certain errors, like inadequate clinical concept extraction, indicate the model's weaknesses. Other errors, such as the difficulty in selecting the base DRG, likely stem from the intricacies of the DRG assignment rules. Furthermore, errors such as the unavailability of correct information required for DRG prediction underscore the limitations of solely relying on discharge summaries for DRG predictions. 

\begin{table*}[!ht]
\small
\captionsetup{font={bf}}
\caption{Example of incorrect DRG predictions}

\label{Table:error analysis}
 \begin{tabular}{l p{7cm}p{2.5cm}p{2.5cm} p{2.5cm} } 
 \toprule
Case ID & Pertinent narratives in discharge summary & True DRG & Predicted DRG & Comment \\ 
\midrule
   Case 1 & altered mental status...respiratory failure...acute blood loss anemia and anemia of chronic disease...clostridium difficile infection...hypotension...was initially on levophed and dopamine... & Heart failure and shock with mcc & Respiratory system diagnosis with ventilator support \>96 hours & Difficulty in selecting base DRG\\\\
   Case 2 & gastrointestinal bleeding...most likely ischemic colitis...viral gastroenteritis...acute renal failure...anemia... & Renal failure with cc & Other digestive system diagnoses with cc & Difficulty in selecting base DRG\\\\
   Case 3 & worsening diabetic foot ulcer...diabetic foot infection...svt...cardiology was consulted... & Cellulitis without mcc & Diabetes with cc & Inadequate clinical concept extraction \\\\
   Case 4 & neutropenic fevers...infectious workup was negative except for a urine culture growing enterococcus...pt is neutropenic, thrombocytopenic, and anemic...hiv-stable... & Kidney and urinary tract infections without mcc & Major hematological and immunological diagnoses except sickle cell crisis and coagulation disorders with mcc & Difficulty in selecting base DRG \\\\
   Case 5 & reported chest pain...soliatry episode of nsvt..ua without pyuria...safe for d/c home... & Esophagitis gastroenteristis and miscellaneous digestive disorders without mcc & Cardiac arrhythmia and conduction disorders with cc & Correct information needed for DRG prediction not available \\\\
   Case 6 & septic arthritis, likely seeded by her recurrence of her e. coli bacteremia...rheum and id recommend wash out...wash out was deferred by orthopedics... & Septicemia or severe sepsis without mv \>96 hours with mcc & Revision of hip or knee replacement with mcc & Inadequate clinical concept extraction \\\\
   Case 7 & acute to subacute hyponatremia...admitted with low na 120...uti with evidence of pyuria... & Kidney and urinary tract infections without mcc & Renal failure with cc & Inadequate clinical concept extraction \\\\
   Case 8 & presents with diffuse acute-on-chronic abdominal pain...gi bleed...treated with octreotide drip and pantoprazole iv...capsule endoscopy was performed...encephalopathy...visual hallucinations... & Septicemia or severe sepsis without mv \>96 hours with mcc & G.i. hemorrhage with cc & Possible incorrect DRG label \\\\
   Case 9 & admitted for altered mental status...delirium....silent aspiration for which received a peg tube...hypertension treated with amlodipine...osa & Esophagitis gastroenteristis and miscellaneous digestive disorders without mcc & Esophagitis gastroenteritis and miscellaneous digestive disorders with mcc & Erroneous cc/mcc \\\\
   Case 10 & presents with word finding difficulties and lethargy...eeg showed moderate encephalopathy...ams was likely due to overmedication...followed by psychiatry - seroquel and abilify were held... & Psychoses & Other disorders of nervous system with cc & Inadequate clinical concept extraction \\\\

 \bottomrule
 \end{tabular}
 We manually reviewed 10 cases for error analysis. For each case, we extracted most pertinent medical problems and their narratives from discharge summaries. Certain errors, like inadequate clinical concept extraction, indicate the model's weaknesses. Other errors, such as the difficulty in selecting the base DRG, likely stem from the intricacies of the DRG assignment rules. Furthermore, errors such as the unavailability of correct information required for DRG prediction underscore the limitations of solely relying on discharge summaries for DRG predictions.   
 
\end{table*}

\begin{figure*}[ht]
    \centering
    \includegraphics[width=.75\textwidth]{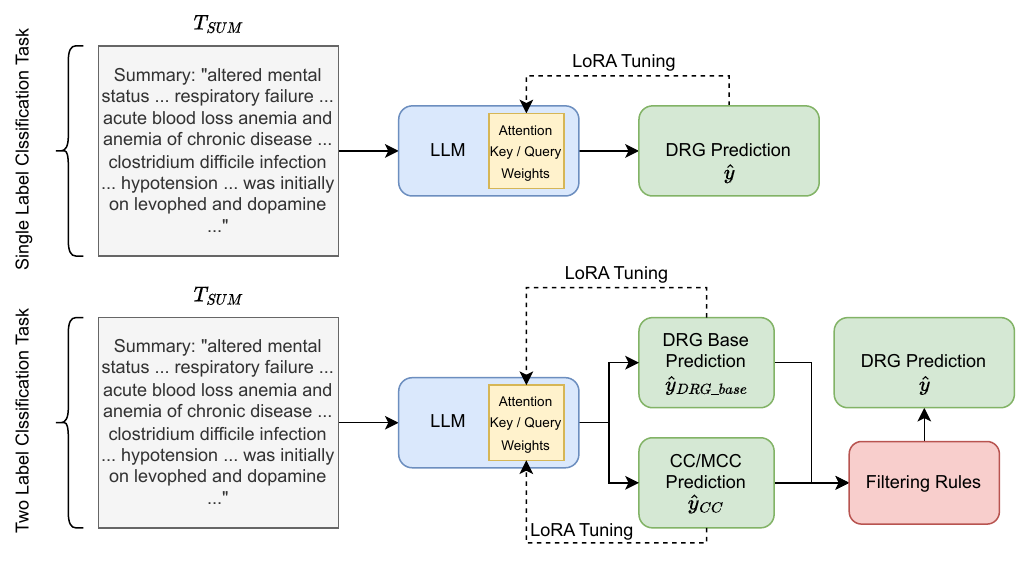}
    \caption{An illustration of both approaches we tested: Single Label Prediction--which directly predicts the DRG code from the text--as well as Two Label Prediction--which breaks down the classification task into 2 tasks. The two predictions are then combined using filtering rules (discovered from data for each DRG) at inference time for the final DRG prediction. LoRA training is used to train the LLM due to computational constraints.}
    \label{fig:method}
\end{figure*}

\section{Discussion} 

\noindent{\bf Large language model context:} Language models based on the transformer architecture, either pretrained or fine-tuned using biomedical corpora, have demonstrated efficacy across a spectrum of NLP benchmarks within the biomedical realm \cite{lee2020biobert, huang2020clinicalbert, gu2021domain}. 
When contrasted with their predecessors rooted in the BERT architecture \cite{devlin2018bert}, LLMs stand out due to their substantial size and their pretraining on expansive, cross-disciplinary text corpora. 
LLMs exhibit a notable capacity for comprehending and reasoning with clinical knowledge.
Without domain-specific fine-tuning or specialized prompt crafting, GPT-4 exceeded the passing score on USMLE by over 20 points and set a new state-of-the-art \cite{nori2023capabilities}. 
On this premise, it is plausible to speculate that once attuned to the medical domain, an LLM could deliver robust performance across diverse NLP tasks, including the prediction of DRGs.

Toward deploying a local LLM, we used LLaMA, a robust and openly accessible foundational LLM with parameters ranging from 7 billion to 65 billion \cite{touvron2023llama}.
Instruction-following models fine-tuned from LLaMA such as Alpaca \cite{alpaca} and Vicuna \cite{vicuna2023}, exhibit performance on par with GPT-3.5. 
Within the medical context, several groups have directed their efforts towards fine-tuning LLaMA. 
Notable examples among these are ChatDoctor (trained on authentic patient-physician dialogues), HuaTuo (fine-tuned with a Chinese medical knowledge graph), and PMC-LLaMA (fine-tuned on biomedical academic papers) \cite{wang2023huatuo, li2023chatdoctor, wu2023pmcllama}. 
These LLaMA-based models focused on medical question answering, yielding encouraging outcomes.

\noindent{\bf Impact of DRG prediction:} In this study, we demonstrated superior performance of the fine-tuned LLaMA in the text classification task of DRG prediction. 
Previous studies have underscored the effectiveness of employing diverse machine learning algorithms and deep neural networks for DRG prediction within healthcare systems outside the United States \cite{gartner2015machine, islam2021deepdrg}. 
These studies focused on using structured data as input variables instead of clinical text. 
More recently, CAML model exhibited superior ability to predict DRGs \cite{liu2021early}. 
CAML model, exclusively utilizing clinical notes, surpassed the performance of a Long Short-Term Memory (LSTM) model using structured clinical variables \cite{liu2021early}. 
When compared with ClinicalBERT, CAML provided improved F1 scores but lower AUC \cite{liu2021early, alsentzer-etal-2019-publicly}. 
We observed that \mname outperformed prior leading models of ClinicalBERT and CAML.

\noindent{\bf Remarks on DRG prediction results:} ClinicalBERT and CAML already stand as robust baselines, with the added benefit of much faster training times (supplement Table 1). While BERT-based models have a maximum input length of 512 tokens, CAML has the flexibility to handle longer context \cite{devlin2018bert, liu2021early}. 
We also observed that the performance of \mname enhanced with the utilization of larger models and longer input context length. 
Interestingly, a recent study revealed that the optimal performance of LLMs is attained when pertinent information is positioned at either the beginning or the end of the input context, with a decline as the input context expands \cite{liu2023lost}. 
In our constrained experiments conducted with a maximum input token limit up to 1024, we have yet to encounter this limitation.
In our study, the performance of both the baseline models and \mname surpassed the outcomes reported in prior research \cite{liu2021early}. 
Beyond the substantially larger training dataset employed in MIMIC-IV compared to MIMIC-III (236,192 vs. 17,815), it is plausible that this enhanced performance is predominantly linked to our strategic input data selection. 

The study by \cite{liu2021early} included only clinical notes charted up to 48 hours post-admission or 48 hours after ICU admission. 
In the MIMIC-III database, a large portion of records during this time window comprises nursing and radiology notes, potentially lacking the pivotal admission History of Present Illness (HPI) notes. 
In contrast, our methodology entailed the utilization of discharge summaries as the input data source. 
Discharge summary is a comprehensive clinical narrative encapsulating pivotal events, diagnostics, and treatments during hospitalization. 
To accommodate the input token limitations of LLaMA, we exclusively focused on the ``brief hospital course'' section of the summary, intentionally excluding other segments such as physical examinations, radiology, laboratory, and medication list. 
Additionally, to enhance data consistency, we formulated an algorithm aimed at addressing discrepancies in DRG nomenclature and assignments across different years.

\noindent{\bf Nuance of DRG prediction task:} In the context of the DRG system, a DRG code comprises a base DRG and a CC/MCC status.
The base DRG represents the principal diagnosis (for medical cases) or procedures (for surgical cases) leading to the patient's admission. 
Meanwhile, CC/MCC categorizations gauge the severity of the patient's condition. 
In the 34.0 version of the MS-DRG system, there are 154 three-way split DRGs, 44 two-way split DRGs with MCC/CC and no CC, 65 two-way split DRGs with MCC and CC/no CC, and 77 base DRGs with no splits (examples in Supplementary Note 1) \cite{centers2016icd}.
We experimented to resemble this structure through a two-label DRG prediction strategy. 
Surprisingly, the top-1 accuracy for CC/MCC stands at 67.5\%, similar to 67.8\% of the base DRG despite the considerably smaller label count (5 labels in CC/MCC vs. 340 labels in base DRG). 
These unexpected results likely stem from the noisy nature of CC/MCC assignment. 
For instance, the DRG code ``pulmonary edema and respiratory failure'' does not have a CC/MCC split. 
Therefore, a hospital stay with this DRG code may truly contain MCC, but the MCC would not be labeled as positive in the training set.
To address this challenge, we formulated rules in both the DRGs dissection phase (extracting base DRGs and CC/MCC from DRGs) and the inference phase (deriving DRGs based on base DRGs and CC/MCC). 
These rules cater to various split scenarios, thus improving accuracy. 
Implementing such rules has culminated in a final DRG prediction accuracy close to single-label prediction (51.5\% vs. 52.0\%).

\noindent{\bf Remarks on error analysis:} Our error analysis also revealed intriguing observations. 
While certain vulnerabilities (e.g., erroneous CC/MCC classification and inadequate clinical concept extraction) present opportunities that theoretically can be addressed through employment of larger LLM and more data, other challenges likely stem from inherent limitations within our training data setup. 
For instance, in Case 2 in Table \ref{Table:error analysis}, despite the discharge summary providing a more comprehensive discussion on gastrointestinal bleeding compared to acute renal failure, the latter was deemed the correct base DRG. 
This selection is guided by the DRG assignment rule, a factor extending beyond the scope of what is directly evident within the discharge summary. 

\noindent{\bf Limitations of our work:} Our study has several limitations. 
1) We were limited by the constraints of the MIMIC-IV dataset and could only use discharge summaries as input data, which are only available after the patient is discharged from the hospital. However, an effective alternative for predicting early DRGs would be to utilize HPI notes and/or Emergency Department (ED) notes. This approach has the potential to significantly impact hospital operations. The ``assessment and plan" in HPI notes are similar in structure to the ``brief hospital course" in discharge summaries. Thus, LLMs might find it easier to extract information related to the principal diagnosis from these notes, given their earlier time stamp in the hospitalization process.

2) We were also restricted by computational resource limitations, so we could only experiment with the LLaMA model up to a parameter size of 13 billion. Unfortunately, we couldn't perform an extensive hyperparameter search. The largest LLaMA models have over 65 billion parameters.

\noindent{\bf Conclusion and future work:} The results presented in this study highlight the potential of adapting LLMs for medical purposes, particularly in predicting DRGs. Future research should involve collaborating with healthcare systems and utilizing admission notes to enable early DRG prediction. Additionally, our findings suggest that experiments utilizing the latest LLMs, including the recently launched 70-billion-parameter LLaMA-2 model with a maximum context length of 4096 tokens \cite{touvron2023llama2}, should be considered. Finally, a crucial area for exploration concerns the practical implications of such DRG prediction, particularly when integrated into existing hospital coding workflows.

\section{Methodology}

\subsection{Dataset and Preprocessing}
We conducted a study using the publicly available MIMIC-IV dataset, which comprises 431,231 unique hospital admissions from 299,712 patients admitted to an ICU or the ED of the Beth Israel Deaconess Medical Center in Boston, Massachusetts \cite{johnson2023mimic}. The dataset covers the period from 2008 to 2019. We used regular expressions to extract the ``brief hospital course" section from the discharge summary as input text. We then filtered the discharge summaries that were of low quality, identified by either duplicated content or containing less than 40 words.

Our focus was on hospitalizations with MS-DRGs. However, Centers for Medicare \& Medicaid Service adjusts MS-DRG regulations annually, resulting in varying DRG assignments for identical conditions over time within the MIMIC-IV dataset \cite{mimiciv-drg}. To address this discrepancy, we designed an algorithm based on clinical knowledge to harmonize MS-DRG codes across different time points to a unified version (Supplementary Method 1). We selected MS-DRG version 34.0 published in 2016, which included a total of 757 DRG codes, 738 of which were present in our dataset \cite{centers2016icd}. We allocated 90\% of the data to the training set and the remaining 10\% to the testing set, stratified by DRG codes.

\subsection{Model Development}
We performed fine-tuning of the LLaMA model using discharge summaries and DRG codes within the context of a classification task. Our approach includes two distinctive strategies (Also shown in Figure~\ref{fig:method}).

\subsubsection{Single label approach}
In this approach, the model generates a single-label multi-class prediction for the DRG code from a training set of natural text discharge summaries $T_{SUM}$ and labels containing $(T_{SUM,i}, y_i) \in \bm{\mathcal{D}}$ \footnote{We omit the index notation $i$ for the rest of the descriptions without loss of generality}. 
First, let us tokenize $T_{SUM}$ based on the LlaMA Tokenzizer into $\bm{K} = tokenize(T_SUM)$. $\bm{K}$ is a list of indices that index into learnable embedding weights. Let $LLM()$ be a function that outputs the embedding for each token after running the transformer model. Finally, the raw logits are calculated as 
$$\bm{\hat{y}} = LLM(\bm{K})_{-1}$$
where we use the last token embedding of $LLM(\bm{K})$ as the predicted raw logit score of each DRG code $\bm{\hat{y}} \in \mathbb{R}^{738}$. Note that this logit score is the raw, unnormalized output of the last layer of the LLM. Before applying the activation function like the softmax function, which converts these scores to probabilities, the values produced by the network are referred to as logits.

The conventional categorical cross-entropy loss function for multi-class classification is used. i.e., a classic multi-class problem with loss:
The target DRG $y$ is an integer between $0$ and $737$ (note that we use an integer representing a specific DRG code for simplicity). 
$$\ell(\bm{\hat{y}}, y) = - \log \frac{\exp(\bm{\hat{y}}_{y})}{\sum_{c=1}^C \exp(\bm{\hat{y}}_{c})}
$$
Where $y \in \{0,1,\dots, 737\}$ is the target DRG, and $\bm{\hat{y}}_c$ is the $c^{th}$ index of $\bm{\hat{y}}$.

\subsubsection{Two-label approach}
In contrast, the two-label approach entails the model initially predicting the base DRG and the CC/MCC status as two separate classification tasks. Subsequently, a mapping rule is applied to derive DRG code. Details on the dissection and inference process from DRGs to base DRGs and CC/MCC status and vice versa can be found in Supplementary Method 2. This approach entailed a loss function configured as the cross-entropy loss of the base DRG, plus half of the cross-entropy loss of the CC/MCC status.

More formally,
\begin{align*}
    & \ell(\bm{\hat{y}},y) = \ell_{DRG\_base}(\bm{\hat{y}}_{DRG\_base}, y_{DRG\_base}) + 
    \\
    & \lambda \ell_{CC}(\bm{\hat{y}}_{CC}, y_{CC})
\end{align*}
Where $\ell_{DRG\_base}(\bm{\hat{y}}_{DRG\_base}, y_{DRG\_base})$ and $\ell_{CC}(\bm{\hat{y}}_{CC}, y_{CC})$ are also categorical cross entropy losses. We chose $\lambda=\frac{1}{2}$ for our work. As shown in Table \ref{Table:two label}, $y_{DRG\_base} \in \{0,1,\dots, 339\}$ and $y_{CC} \in \{0,\dots,4\}$, representing the categories of [``without CC/MCC'', ``with CC'', ``with MCC'', ``without MCC'', and ``not applicable''] respectively. 

To enable ease of implementation, we used an output logit dimension of $\bm{\hat{y}}\in \mathbb{R}^{340+5}$ and indexed the first 340 dimensions for $\bm{\hat{y}}_{DRG\_base} = \bm{\hat{y}}_{0,\dots,339}$ and indexed the last 5 dimensions for $\bm{\hat{y}}_{CC} = \bm{\hat{y}}_{340,\dots,344}$. At inference time, we take the base DRG and CC/MCC predictions as the argmax of their respective logits. 
\begin{align*}
& \hat{y}_{DRG\_base} = argmax_{\hat{y}_{DRG\_base}} ( \bm{\hat{y}}_{DRG\_base} ) \\
& \hat{y}_{CC} = argmax_{\hat{y}_{CC}} ( \bm{\hat{y}}_{CC} | \hat{y}_{CC} \in V_{\hat{y}_{DRG\_base}}) \\
\end{align*}
Subsequently, we apply the mapping rule, as detailed in Supplementary Method 2, to derive the final DRG prediction from base DRG and CC/MCC labels. 


\subsubsection{Addressing Computational Constraints via LoRA Training}
Given the constraints of available computational resources, an extensive hyperparameter search was not viable. 
Instead, our focus encompassed exploring the performance across diverse model sizes and token lengths. 
We used LoRA during training, which involves freezing the pre-trained model weights and incorporating trainable rank decomposition matrices into each layer of the transformer architecture \cite{hu2021lora}. Lora training of the attention mechanism is shown in Figure~\ref{fig:method}.

As a quick summary, let us assume that we have original weight matrix $\bm{W_0} \in \mathbb{R}^{d \times k}$. LoRA works by adding a low-rank matrix to the original weight matrix: $\Delta \bm{W} + \bm{W_0}, \Delta \bm{W}  = \bm{B}\bm{A}$ where $\bm{B}\in \mathbb{R}^{d \times r}$ and $\bm{A}\in \mathbb{R}^{r\times k}$ . Note that one should choose $r \ll \min(d, k)$ and only adapt the attention weights to ensure constraints on the dimensionality of the new weights and preserve original model performance. Training is only performed on this $\Delta \bm{W}$, and original model weights are kept the same. We also only tune the weights of the attention mechanism for further cost savings while preserving performance.

\subsubsection{Training Details}
Model training adopted standard Huggingface training framework and the sequence classification module \cite{wolf2020huggingfaces}. 
Since LLaMA is a decoder-only (causal) model, we follow the traditional approach of using the embedding of the last token to do the classification, as other causal models (e.g. GPT-2 \cite{radford2019language}) do.
Logits score of each DRG label was calculated from this linear output layer, and probabilities of DRGs could be derived using a softmax function. 

We referenced the training protocol of Alpaca-Lora \cite{lora-alpaca}. 
Our model was trained using cross-entropy loss with the Adam optimizer (learning rate = \num{2e-5} and weight decay = 0.01) for 3 epochs on all training data and batch size of 4. Lora parameters were configured with r set to 8, an alpha value of 8, and a dropout rate of 0.05. 
All attention blocks were included in the Lora target modules. 
The training regimen for all \mname models were executed on a singular Nvidia RTX A6000 GPU with 48GB of graphics memory.

\subsection{Baseline Models}
As baseline models for benchmarking, We selected CAML \cite{mullenbach2018explainable, liu2021early} and ClinicalBERT \cite{alsentzer-etal-2019-publicly}. CAML is an adjusted convolutional neural network (CNN). In CAML, clinical notes are tokenized and embedded with pre-trained word embeddings to form input representations. Subsequently, inputs are passed on to a neural network with one-dimensional convolutions that pool CNN features using the attention mechanism. In line with the approach detailed in \citet{liu2021early}, our training of CAML included early stop when there was no improvement in micro-averaged F1 score for 10 consecutive epochs, with a maximum epochs of 50. All default hyperparameters were kept, except for max\_seq\_length which was set to 512. 

ClinicalBERT was built upon BioBERT, a domain-specific BERT model pre-trained on PubMed abstracts and full-text articles from PubMed Central \cite{Lee_2019}. ClinicalBERT performed further pre-training of BioBERT using 2 million clinical notes from MIMIC-III \cite{johnson2016mimic}. In our fine-tuning process of ClinicalBERT, we conducted three training epochs, same as \mname. We set a learning rate of \num{2e-5} and a batch size of 16, consistent with previous recommended practice for classification-oriented fine-tuning of BERT \cite{devlin2018bert, adhikari2019docbert}. 

\subsection{Statistical analysis}
We used the implementation from \cite{liu2021early} to calculate AUC and F1-score in both macro- and micro- approach for predictive models. 
We also reported accuracy of DRG prediction at top one, five and ten results. Standard deviations was calculated using a bootstrapping procedure with 30 iterations. 
For each bootstrap iteration, we randomly resampled the whole sample size from the testing set with replacement.
Smoothing spline fit in Figure \ref{fig:2a} was performed using npreg package in R with generalized cross-validation method and default parameters \cite{npreg}.

\subsection{Data availability}
Access to MIMIC-IV can be requested at \url{https://physionet.org/content/mimiciv/},
which requires a signed safe usage agreement.

\subsection{Code availability}
Scripts for this work were written in Python. They are available with accompanied documentation at \url{https://github.com/hanyin88/drg-llama}.

\subsection{Ethical Concerns}
MIMIC-IV is a free EHR dataset that is deidentified according to the Health Insurance Portability and Accountability Act (HIPAA) Safe Harbor provision \cite{johnson2023mimic}

Since we primarily used open source models such as LLaMA and ClinicalBERT from Huggingface, an open source repository of machine learning models \cite{wolf2020huggingfaces} as well as CAML from github, and trained it on MIMIC, privacy risks are quite low. 
However, this risk should not be counted out when working with LLMs, and it is possible that LLaMA and ClinicalBERT may be trained on sensitive data in their respective pretrainining stages.

\subsection{Acknowledgement}
This research was supported by NSF award SCH-2205289, IIS-2034479, SCH-2014438.

\subsection{Author Contributions}
H.W. designed, conducted, and analyzed the results of experiments. H.W. and C.G wrote the original draft. J.S. obtained funding and computing resource for the project. All authors contributed to the conceptualization of the research questions. All authors reviewed, revised, and approved the final manuscript.

\clearpage
\bibliography{reference.bib}
\end{document}


\maketitle

\section{Supplementary Note 1: Examples of DRGs split in MS-DRG version 34.0}
\begin{enumerate}
\item Three-way split DRGs: 
\begin{itemize}
\item DRG 11: tracheostomy for face mouth and neck diagnoses with MCC
\item DRG 12: tracheostomy for face mouth and neck diagnoses with CC
\item DRG 13: tracheostomy for face mouth and neck diagnoses without CC/MCC
\end{itemize}
\item Two-way spit DRGs with MCC/CC and no CC: 
\begin{itemize}
    \item DRG 52: spinal disorders and injuries with CC/MCC
    \item DRG 53: spinal disorders and injuries without CC/MCC
\end{itemize}
\item Two-way split DRGs with MCC and CC/no CC: 
\begin{itemize}
    \item DRG 56: degenerative nervous system disorders with MCC
    \item DRG 57: degenerative nervous system disorders without MCC
\end{itemize}
\item Base DRGs with no splits: 
\begin{itemize}
    \item DRG 69: transient ischemia
\end{itemize}
\end{enumerate}

\section{Supplementary Method 1. Process to address different DRG versions}
We designed an algorithm grounded in clinical knowledge, aimed at unifying MS-DRG codes across different time points to a single version—specifically, MS-DRG version 34.0 published in 2016 \cite{centers2016icd}. The process include:
\begin{enumerate}
\item Standardize use of abbreviations and capitalization within DRGs. For example, we replaced all "W/O" to "WITHOUT", "CATH" to "CATHETERIZATION" and "PROC" to "PROCEDURES".
\item Using a fuzzy string match algorithm (TheFuzz: https://github.com/seatgeek/thefuzz) to find those DRGs not matching to any DRG v.34 codes. 
\item An internal medicine physician manually reviewed all DRG codes from step 2, and manually assigned these codes to the most appropriate MS-DRG v.34.0 codes if applicable. 
\item Of note, after above steps there are several historical DRGs left without appropriate DRG v.34 codes assignment. For example, "URINARY STONES W MCC" and "NASAL TRAUMA AND DEFORMITY WITH CC". These hospitalizations were excluded from the cohort. 
\end{enumerate}

\section{Supplementary Method 2. Process to dissect and derive DRGs to/from base DRGs and CC/MCC}
We first used regular expression to obtain principal diagnosis/procedures in MS-DRG v.34.0, by extracting strings prior to the description of CC/MCC. For example, in DRG 11 of "tacheostomy for face mouth and neck diagnoses with mcc",  the principal diagnosis is "tacheostomy for face mouth and neck diagnoses". After this step, 340 different principal diagnosis/procedures are identified as base DRGs. 

We assigned CC/MCC status to one of the five labels: "without CC/MCC," "with CC," "with MCC," "without MCC," and "not applicable". Of note, an important detail is that if a DRG code does not explicitly describe CC/MCC status, we will assign a label of "not applicable". Such an example is DRG 69 "transient ischemia". We realize such classification might bring in noisy signals for models to learn (as a patient with "transient ischemia" can indeed has CC/MCC), but found it better than assigning to an alternative label such as "without CC/MCC" which would be more erroneous.     

When inferencing DRGs from base DRGs and CC/MCC, we  developed a rule based on logic and clinical knowledge. First, we evaluate whether predicted principal diagnosis/procedure matches target base DRG. Second, if the predicted CC/MCC label is in the CC/MCC set of the target base DRG, we make  comparison directly. Third, for those predicted CC/MCC labels not in the CC/MCC set of the target base DRG, we apply a mapping procedure based on different MS-DRG splits as listed in Supplementary Note 1. For example, if a MS-DRG code has no split, such as DRG 69 "transient ischemia", then any CC/MCC predictions can be mapped to the correct DRG (as long as the base DRG matches). Another example would be MS-DRG 56 and 57, where there are two splits of CC/MCC status ("with MCC" and "without MCC"). In this case we will map predictions of "without CC/MCC", "with CC" and "not applicable" all to the label of "without MCC" for final inference.

\newpage
\section{Supplementary Results}

\subsection{Distribution of training cases per DRG}
The distribution of training cases per DRG is quite imbalanced with a long tail (Supplementary Figure 1). The median training cases per DRG is 124.5, and there are only 12 DRG codes with a training cases exceeding 2000. 

\begin{figure}[h]
\renewcommand{\figurename}{Supplementary Figure}
\includegraphics[width=0.8\textwidth]{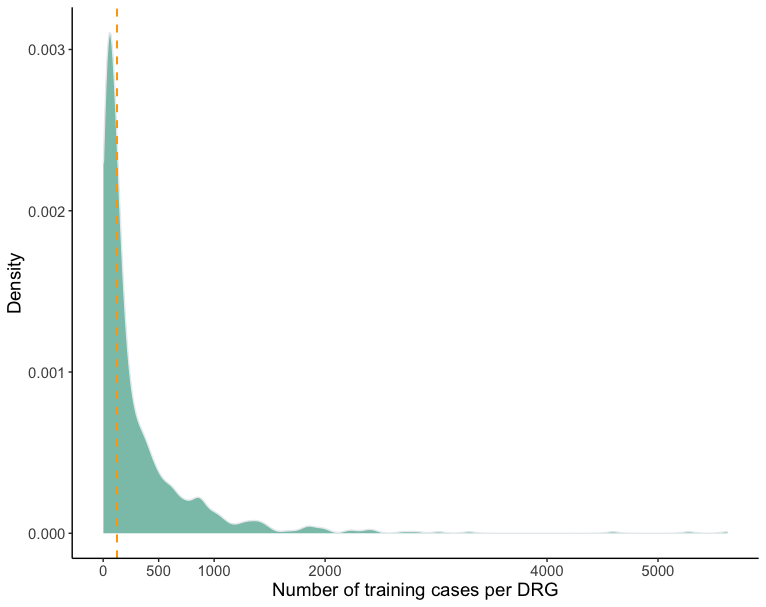}
\captionsetup{labelfont=bf={bf}}
\caption{\textbf{Density plot showing distribution of training cases per DRG.} Density plots are conceptually equivalent to histograms, but use probability density to display where values are concentrated. The dashed line represent the median number of training cases.}
\centering
\end{figure}

\subsection{Computational cost}
The training and evaluation times for all models are presented in Supplementary Table 1. Notably, ClinicalBERT and CAML exhibited considerably shorter duration to train in comparison to \mname. As the maximal input token size doubled, the training time for \mname increased by approximately 90\%. Moreover, when \mname model size expanded from 7 billion to 13 billion parameters, the training time showed an increment of roughly 70\%.

\begin{table*}[ht]
\small
\renewcommand{\tablename}{Supplementary Table}
\captionsetup{font={bf}}
\caption{Computational costs of all models}
\label{Table:8}
 \begin{tabular}{p{3.5cm} p{3.5cm} p{3.5cm} p{3.5cm} } 
 \toprule
Model & Max input token size & Training time (hours) & Evaluation steps per second \\ 
\midrule
     \mname-7B & 340 & 104 & 2.39\\ 
     & 512 & 151 &  1.70 \\
      & 1024 & 285 & 0.84 \\
     \mname-13B & 340 & 177 & 1.54 \\ 
     & 512 & 258 & 1.08 \\
      & 1024 & 498 &  0.52 \\ 
     ClinicalBERT & 512 & 5 & 7.26 \\
     CAML & 512 & 0.8 & 98.53 \\ 
 \bottomrule
\end{tabular}
\\
All training and evaluation were performed on one 48GB Nvidia RTX A6000 GPU. Batch sizes for \mname, ClinicalBERT and CAML were 4, 16 and 32, respectively. 
\end{table*}

\newpage
\subsection{Analysis of DRGs with best and worst prediction performance}
The nine DRGs with a top-1 prediction accuracy of 100\% by \mname are detailed in Supplementary Table 2. Majority of these DRGs (8 out of 9) fall within the category of surgical DRGs, representing infrequent yet highly specialized procedures. Interestingly, the median number of training cases for this particular set of DRGs is low at 23. Supplementary Table 3 and Table 4, on the other hand, outline medical and surgical DRGs with a top-5 prediction accuracy of 0\%, respectively. Within this cohort, the median number of training samples is 17.5 for medical DRGs and 16 for surgical DRGs. Among these challenging-to-predict DRGs, many are clinically and semantically difficult to differentiate from alternative options. For instance, consider the distinction between DRG 671, "Urethral Procedures With CC/MCC," and DRG 668, "Transurethral Procedures With MCC."

\begin{table*}[ht]
\small
\renewcommand{\tablename}{Supplementary Table}
\captionsetup{font={bf}}
\caption{DRGs with top-1 prediction accuracy of 100\% by \mname}
\label{Supplement table: 100 accuracy}
 \begin{tabular}{c l c c} 
 \toprule
MS-DRG Code & Description & Medical or Surgical & No. of training cases \\ 
\midrule
777 & Ectopic Pregnancy & Medical & 59 \\
801 & Splenectomy Without CC/MCC & Surgical & 23 \\
8 & Simultaneous Pancreas and Kidney Transplant & Surgical & 21 \\
10 & Pancreas Transplant & Surgical & 19 \\
799 & Splenectomy With MCC & Surgical & 23 \\
652 & Kidney Transplant & Surgical & 380 \\
116 & Intraocular Procedures With CC/MCC & Surgical & 17 \\
16 & Autologous Bone Marrow Transplant With CC/MCC & Surgical & 155 \\
114 & Orbital Procedures Without CC/MCC & Surgical & 31 \\
 \bottomrule
\end{tabular}
\\
Results from \mname-7B with input token window of 512.

\end{table*}


\begin{table*}[ht]
\small
\renewcommand{\tablename}{Supplementary Table}
\captionsetup{font={bf}}
\caption{Medical DRGs with top-5 prediction accuracy of 0\% by \mname}
\label{Table:6}
 \begin{tabular}{c p{10cm} c} 
 \toprule
MS-DRG Code & Description & No. of training cases \\ 
\midrule
53 & Spinal Disorders and Injuries Without CC/MCC & 17 \\
63 & Acute Ischemic Stroke With Use of Thrombolytic Agent Without CC/MCC & 14 \\
67 & Nonspecific CVA and Precerebral Occlusion Without Infarction With MCC & 17 \\
77 & Hypertensive Encephalopathy With MCC & 21 \\
79 & Hypertensive Encephalopathy Without CC/MCC & 6 \\
80 & Nontraumatic Stupor and Coma With MCC & 33 \\
81 & Nontraumatic Stupor and Coma Without MCC & 78 \\
88 & Concussion With MCC & 12 \\
89 & Concussion With CC & 25 \\
90 & Concussion Without CC/MCC & 31 \\
122 & Acute Major Eye Infections Without CC/MCC & 13 \\
124 & Other Disorders of the Eye With MCC & 35 \\
150 & Epistaxis With MCC & 32 \\
182 & Respiratory Neoplasms Without CC/MCC & 19 \\
284 & Acute Myocardial Infarction Expired With CC & 20 \\
285 & Acute Myocardial Infarction Expired Without CC/MCC & 10 \\
290 & Acute and Subacute Endocarditis Without CC/MCC & 8 \\
297 & Cardiac Arrest Unexplained With CC & 10 \\
383 & Uncomplicated Peptic Ulcer With MCC & 17 \\
533 & Fractures of Femur With MCC & 11 \\
537 & Sprains Strains and Dislocations of Hip Pelvis and Thigh With CC/MCC & 18 \\
538 & Sprains Strains and Dislocations of Hip Pelvis and Thigh Without CC/MCC & 7 \\
548 & Septic Arthritis With MCC & 12 \\
550 & Septic Arthritis Without CC/MCC & 12 \\
594 & Skin Ulcers Without CC/MCC & 15 \\
598 & Malignant Breast Disorders With CC & 43 \\
688 & Kidney and Urinary Tract Neoplasms Without CC/MCC & 6 \\
695 & Kidney and Urinary Tract Signs and Symptoms With MCC & 25 \\
722 & Malignancy Male Reproductive System With MCC & 19 \\
725 & Benign Prostatic Hypertrophy With MCC & 19 \\
730 & Other Male Reproductive System Diagnoses Without CC/MCC & 6 \\
756 & Malignancy Female Reproductive System Without CC/MCC & 8 \\
757 & Infections Female Reproductive System With MCC & 23 \\
836 & Acute Leukemia Without Major O.R. Procedure Without CC/MCC & 26 \\
845 & Other Myeloproliferative Disorders or Poorly Differentiated Neoplastic Diagnoses Without CC/MCC & 16 \\
886 & Behavioral and Developmental Disorders & 30 \\
887 & Other Mental Disorder Diagnoses & 30 \\
913 & Traumatic Injury With MCC & 21 \\
 \bottomrule
\end{tabular}

Results from \mname-7B with input token window of 512.

\end{table*}

\newpage


\begin{table*}[ht]
\small
\renewcommand{\tablename}{Supplementary Table}
\captionsetup{font={bf}}
\caption{Surgical DRGs with top-5 prediction accuracy of 0\% by \mname}
\label{Table:7}
 \begin{tabular}{c p{10cm} c} 
 \toprule
MS-DRG Code & Description & No. of training cases \\ 
\midrule
13 & Tracheostomy for Face Mouth and Neck Diagnoses Without CC/MCC & 13 \\
115 & Extraocular Procedures Except Orbit & 24 \\
117 & Intraocular Procedures Without CC/MCC & 5 \\
130 & Major Head and Neck Procedures Without CC/MCC & 13 \\
135 & Sinus and Mastoid Procedures With CC/MCC & 8 \\
138 & Mouth Procedures Without CC/MCC & 26 \\
231 & Coronary Bypass With PTCA With MCC & 14 \\
232 & Coronary Bypass With PTCA Without MCC & 11 \\
241 & Amputation for Circulatory System Disorders Except Upper Limb and Toe Without CC/MCC & 23 \\
257 & Upper Limb and Toe Amputation for Circulatory System Disorders Without CC/MCC & 22 \\
258 & Cardiac Pacemaker Device Replacement With MCC & 16 \\
263 & Vein Ligation and Stripping & 16 \\
332 & Rectal Resection With MCC & 26 \\
341 & Appendectomy Without Complicated Principal Diagnosis With MCC & 16 \\
344 & Minor Small and Large Bowel Procedures With MCC & 50 \\
347 & Anal and Stomal Procedures With MCC & 30 \\
410 & Biliary Tract Procedures Except Only Cholecyst With or Without C.D.E. Without CC/MCC & 35 \\
411 & Cholecystectomy With C.D.E. With MCC & 7 \\
412 & Cholecystectomy With C.D.E. With CC & 19 \\
413 & Cholecystectomy With C.D.E. Without CC/MCC & 14 \\
422 & Hepatobiliary Diagnostic Procedures Without CC/MCC & 18 \\
425 & Other Hepatobiliary or Pancreas O.R. Procedures Without CC/MCC & 14 \\
458 & Spinal Fusion Except Cervical With Spinal Curvature or Malignancy or Infection or Extensive Fusions Without CC/MCC & 9 \\
487 & Knee Procedures With PDX of Infection Without CC/MCC & 25 \\
507 & Major Shoulder or Elbow Joint Procedures With CC/MCC & 17 \\
518 & Back and Neck Procedures Except Spinal Fusion With MCC or Disc Device or Neurostimulator & 35 \\
576 & Skin Graft Except for Skin Ulcer or Cellulitis With MCC & 14 \\
624 & Skin Grafts and Wound Debridement for Endocrine Nutritional and Metabolic Disorders Without CC/MCC & 13 \\
630 & Other Endocrine Nutritional and Metabolic O.R. Procedures Without CC/MCC & 22 \\
662 & Minor Bladder Procedures With MCC & 7 \\
666 & Prostatectomy With CC & 10 \\
671 & Urethral Procedures With CC/MCC & 14 \\
711 & Testes Procedures With CC/MCC & 12 \\
715 & Other Male Reproductive System O.R. Procedures for Malignancy With CC/MCC & 9 \\
716 & Other Male Reproductive System O.R. Procedures for Malignancy Without CC/MCC & 8 \\
802 & Other O.R. Procedures of the Blood and Blood Forming Organs With MCC & 31 \\
829 & Myeloproliferative Disorders or Poorly Differentiated Neoplasms With Other O.R. Procedure With CC/MCC & 68 \\
830 & Myeloproliferative Disorders or Poorly Differentiated Neoplasms With Other O.R. Procedure Without CC/MCC & 17 \\
855 & Infectious and Parasitic Diseases With O.R. Procedure Without CC/MCC & 13 \\
901 & Wound Debridements for Injuries With MCC & 23 \\
903 & Wound Debridements for Injuries Without CC/MCC & 32 \\
928 & Full Thickness Burn With Skin Graft or Inhalation Injury With CC/MCC & 6 \\
939 & O.R. Procedures With Diagnoses of Other Contact With Health Services With MCC & 29 \\
959 & Other O.R. Procedures for Multiple Significant Trauma Without CC/MCC & 55 \\
970 & HIV With Extensive O.R. Procedure Without MCC & 10 \\
 \bottomrule
\end{tabular}

Results from \mname-7B with input token window of 512.

\end{table*}

\clearpage
\bibliographystyle{aaai22}
\bibliography{reference.bib}